\documentclass{ecai} 

\usepackage{latexsym}
\usepackage{amssymb}
\usepackage{amsmath}
\usepackage{amsthm}
\usepackage{booktabs}
\usepackage{enumitem}
\usepackage{graphicx}
\usepackage{color}

\usepackage{multirow}
\usepackage{subcaption}
\usepackage{algorithmic}
\usepackage{textcomp} 
\usepackage{float}
\usepackage{titlesec}
\usepackage{caption}

\newcommand{\BibTeX}{B\kern-.05em{\sc i\kern-.025em b}\kern-.08em\TeX}

\begin{document}
\begin{frontmatter}

\paperid{123} 


\title{PuriLight: A Lightweight Shuffle and Purification Framework for Monocular Depth Estimation}

\author[A]{\fnms{Yujie}~\snm{Chen}}
\author[A]{\fnms{Li}~\snm{Zhang}\thanks{Corresponding authors.}}
\author[B]{\fnms{Xiaomeng}~\snm{Chu}\protect\footnotemark[*]} 
\author[A]{\fnms{Tian}~\snm{Zhang}}

\address[A]{Department of Mathematics, Hefei University of Technology, Hefei, China}
\address[B]{Department of Computer Science and Technology, University of Science and Technology of China, Hefei, China}




\begin{abstract}
We propose PuriLight, a lightweight and efficient framework for self-supervised monocular depth estimation, to address the dual challenges of computational efficiency and detail preservation. While recent advances in self-supervised depth estimation have reduced reliance on ground truth supervision, existing approaches remain constrained by either bulky architectures compromising practicality or lightweight models sacrificing structural precision. These dual limitations underscore the critical need to develop lightweight yet structurally precise architectures. Our framework addresses these limitations through a three-stage architecture incorporating three novel modules: the Shuffle-Dilation Convolution (SDC) module for local feature extraction, the Rotation-Adaptive Kernel Attention (RAKA) module for hierarchical feature enhancement, and the Deep Frequency Signal Purification (DFSP) module for global feature purification. Through effective collaboration, these modules enable PuriLight to achieve both lightweight and accurate feature extraction and processing. Extensive experiments demonstrate that PuriLight achieves state-of-the-art performance with minimal training parameters while maintaining exceptional computational efficiency. Codes will be available at https://github.com/ishrouder/PuriLight.  
\end{abstract}

\end{frontmatter}


\section{Introduction}	
Numerous applications in robotics, Unmanned Aerial Vehicles (UAVs), and Virtual Reality (VR) fundamentally depend on precise depth perception. Monocular depth estimation, the task of inferring depth information from a single image, has emerged as a foundational technique for reconstructing 3D scene geometry from individual images. As many real-world scenarios lack large-scale, dense ground truth depth required for supervised learning \cite{eigen2014depth, chang2018}, self-supervised methods \cite{godard2019digging, lee2024robust, wei2023surrounddepth, liu2024mono} have gained prominence by leveraging geometric constraints from stereo image pairs \cite{garg2016, godard2017} or temporal sequences in monocular videos \cite{zhou2017un}. 
Recent advances in monocular depth estimation have primarily focused on achieving architectural innovations. Early works concentrated on designing new CNN architectures \cite{eigen2014depth, laina2016deeper, fu2018deep, yin2018geonet}. However, it is well-known that convolutional operations in CNNs possess local receptive fields, limiting their ability to capture global information from distant regions. Fortunately, the introduction of ResNet \cite{he2016deep} enabled the training of deeper network models. Consequently, subsequent research efforts have been directed toward developing more sophisticated architectures and deeper backbone networks \cite{godard2019digging, lyu2021hr, zhou2021self}, achieving improved performance at the cost of increased model complexity. Current mainstream monocular depth estimation methods, such as those proposed in \cite{yang2021trans, ran221vis} based on Vision Transformers (ViT) and \cite{ke2024rep} utilizing Stable Diffusion, primarily focus on employing deeper networks or more complex architectures to achieve superior performance. However, due to their extensive training parameters and high computational complexity, these methods exhibit slow inference speeds and impose substantial demands on computational resources, thereby posing challenges for deployment on edge devices while facilitating the construction of lightweight structures.

\begin{figure}[H] 
\centering 
\includegraphics[width=0.48\textwidth]{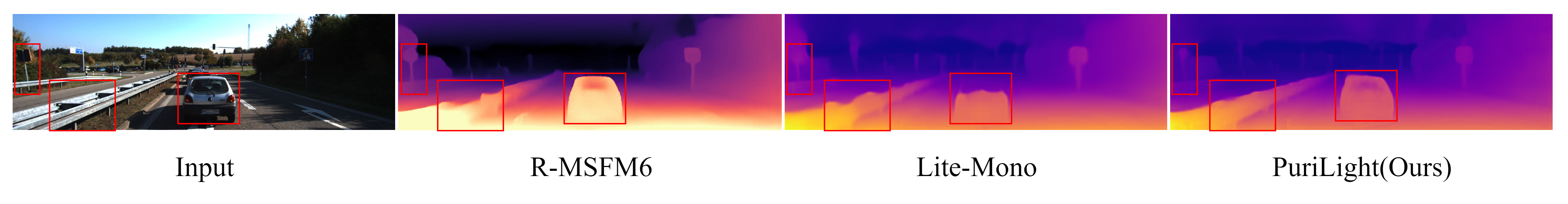} 
\captionsetup{justification=centering}
\caption{The proposed PuriLight delivers more refined depth details and higher accuracy with minimal parameters compared to other representative lightweight methods.}
\label{fig:example} 
\end{figure}

The pursuit of lightweight solutions has inspired multiple architectural innovations. Zhao et al. \cite{rmsfm} developed a multi-scale feature modulation module by truncating the last two blocks of ResNet-18, effectively reducing parameters. However, this approach exhibited low receptive fields; additionally, it suffered from high computational complexity and slow inference speeds due to iterative processing. Zhang et al. \cite{litemono} proposed a hybrid CNN-Transformer architecture to balance computational efficiency and performance. While this framework successfully reduced computational overhead, it demonstrated suboptimal performance in detail-sensitive regions. This limitation manifested particularly as blurred object boundaries in depth estimation results, which are critical for model performance evaluation. 

This paper proposes a novel and efficient lightweight deep learning architecture, PuriLight, to enrich monocular depth estimation frameworks. Unlike previous lightweight methods relying on deep networks, our approach decreases network depth while expanding feature channel dimensions. This enables learning of more accurate contour details. Meanwhile, the method utilizes frequency domain characteristics to emphasize enhancing the extraction of image contours and structural details, achieving a more precise depth map. Three innovative modules are primarily designed: the Shuffle-Dilation Convolution (SDC) module, the Rotation-Adaptive Kernel Attention (RAKA) module, and the Deep Frequency Signal Purification (DFSP) module. The SDC module efficiently captures multi-scale local features. Subsequently, the RAKA module receives and enhances these enriched multi-scale features. Finally, the DFSP module performs signal purification operations on the enhanced features. The contributions of this paper can be summarized as follows:

\begin{itemize}
\item We design a new lightweight and efficient network architecture for self-supervised monocular depth estimation, which is well suited for resource-constrained scenarios.

\item This work introduces three novel modules, namely SDC, RAKA, and DFSP, which focus on capturing more precise structural features and contour details, critical for depth estimation tasks, while allowing a reduction in network depth.

\item Compared with other highly competitive models on the KITTI \cite{kitti} dataset, the architecture uses the least number of parameters and achieves the most advanced level of precision. The inference speed of the model is validated across diverse computing platforms, demonstrating the superiority of PuriLight. Experiments on the Make3D \cite{make3D} dataset show that the model has excellent generalizability and is suitable for most real-world scenarios. 
\end{itemize}


\section{Related Work}
\label{sec:Related Work}
\subsection{Supervised Monocular Depth Estimation}
Supervised monocular depth estimation networks utilize LiDAR-acquired ground truth depth data as supervision signals, formulating the task as a depth regression problem. These networks extract features from input images and learn the mapping relationships between RGB values and depth. Eigen et al. \cite{eigen2014depth} pioneered deep learning-based monocular depth estimation, establishing new state-of-the-art results on NYU Depth \cite{silb2012} and KITTI \cite{kitti} datasets. Subsequently, following the breakthrough success of ResNet \cite{he2016deep} in image recognition, Laina et al. \cite{laina16deep} introduced its adaptation to depth estimation. Then, Fu et al. \cite{fu2018deep} innovatively redefined depth estimation as an ordinal regression task. Patni et al. \cite{patni2024} effectively utilized RGB images as conditional guidance to steer the diffusion process. Although supervised models achieve high accuracy, they face critical drawbacks: reliance on expensive LiDAR for dense depth annotations and vulnerability to sparse, noisy data that undermines generalization. These limitations severely constrain their robustness and deployment in complex real-world scenarios. 

\subsection{Self-Supervised Monocular Depth Estimation}
The challenge of self-supervised monocular depth estimation in the absence of dense ground truth supervision has spurred significant methodological innovations. Garg et al. \cite{garg2016} pioneered a paradigm shift by reformulating depth estimation as a view synthesis problem. Then Godard et al. \cite{godard2017} introduced critical enhancements through a loss of consistency of disparity between left-right predictions. To address inherent limitations in dynamic scenes involving occlusions and moving objects, Zhou et al. \cite{zhou2017uns} proposed a groundbreaking joint training architecture. Their framework simultaneously optimized a depth prediction network and a pose estimation network across consecutive monocular video frames. Godard et al. \cite{godard2019digging} later proposed the Monodepth2 framework, demonstrated through empirical evidence that systematic loss function redesign could surpass performance gains from architectural complexity. This principled approach not only simplified network architecture but also improved depth estimation accuracy, particularly in edge-aware detail preservation.

\begin{figure*}[htbp] 
  \centering 
  \includegraphics[width=0.98\textwidth]{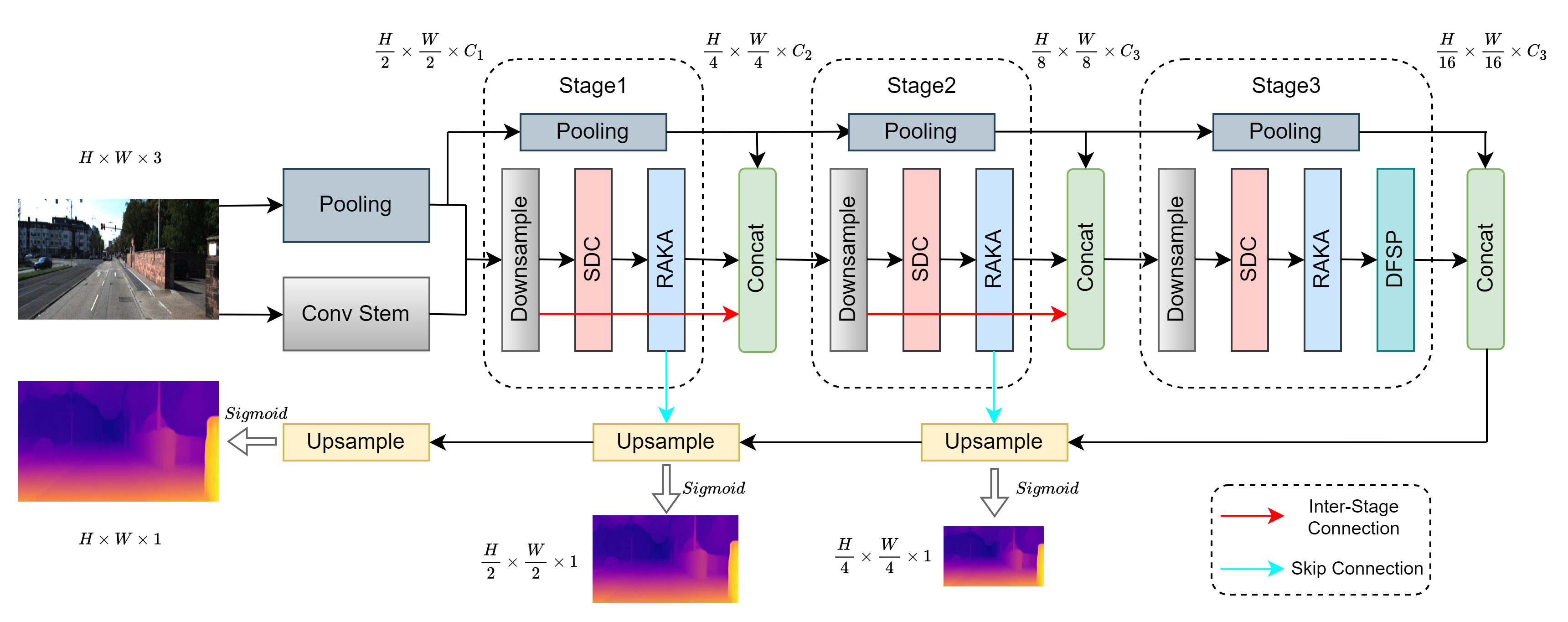} 
  \vspace{0.5cm}
  \captionsetup{justification=centering}
  \caption{Overall Architecture of the PuriLight. The proposed architecture comprises three innovative modules: SDC, RAKA, and DFSP, with detailed implementations illustrated in Figure {\color{red} 3}. }
  \label{fig:example2} 
  \vspace{0.5cm}
\end{figure*}

\section{Methods}
\label{sec:methods}
This paper proposes an efficient and lightweight encoder capable of extracting multi-scale global and local features while adaptively preserving critical features. Our method emphasizes the flexible utilization of shuffle and purification operations to sufficiently enhance inter-channel information exchange and retain essential characteristics. The approach achieves clearer structural details and object contours while maintaining a lightweight architecture. We introduce system overview in Section {\color{red} 3.1}. The architecture of the proposed depth estimation network, PuriLight, is shown in Figure {\color{red}2}, where the encoder design (Sections {\color{red}3.2} and {\color{red}3.3}) and decoder implementation details (Section {\color{red}3.4}) are elaborated in their respective sections. Image inputs are processed by PuriLight to generate multi-scale depth maps, while adjacent frames are fed into PoseNet (Section {\color{red}3.5}) to estimate camera motion.

\subsection{System Overview}

As illustrated in Figure {\color{red}2}, PuriLight consists of three stages, each incorporating pooling layers and inter-stage connections to assist feature representation. It initially processes an input image of size ${H \times W \times 3}$ through two sequential operations: dual ${3 \times 3}$  convolutions followed by pooling layers. Then, features are processed through a sequential pipeline comprising three dedicated modules: SDC, RAKA, and DFSP. Notably, the DFSP module is deployed only in the final processing stage to address accumulated cross-domain loss while maintaining computational efficiency. The hierarchical architecture generates multi-scale features at 1/4, 1/8, and 1/16 the input resolution, respectively.

\textbf{Design and analysis.} (i) It is widely recognized that shallow convolutional neural networks suffer from limited receptive fields, which fundamentally restricts their capacity to capture long-range contextual information. Moreover, given the inherent constraints of lightweight architectures, the number of available feature channels tends to be substantially reduced. To address these dual limitations, we propose a novel Shuffle-Dilation Convolution (SDC) module that synergistically integrates dilated convolution \cite{yu2016multi} with channel shuffle operations \cite{zhang2018shu}. The SDC module expands the network's receptive field without introducing additional parameters, making it particularly suitable for resource-constrained architectures. Second, the channel shuffle mechanism enables more efficient utilization of feature channels by routing half of the processed channels to subsequent modules, achieving effective and lightweight design requirements.

(ii) To further enhance local features, we introduce a Rotation-Adaptive Kernel Attention (RAKA) module. Inspired by \cite{triplet}, our improved implementation addresses critical efficiency concerns in monocular depth estimation tasks. The original architecture's high computational overhead (as measured in FLOPs) made it impractical for lightweight applications. Our improved RAKA module processes multi-scale features from the SDC output through three orthogonal dimensions, employing a combination of pooling layers and adaptive convolutional layers. This optimized configuration achieves superior feature enhancement with little computational overhead.

(iii) In lightweight monocular depth estimation tasks, global feature processing constitutes a critical component. Previous approaches have primarily focused on employing Multi-Head Self-Attention (MHSA) modules from Transformer \cite{transformer} and its variants. However, the quadratic computational complexity of these modules with respect to input dimension poses significant constraints when designing lightweight network architectures. On the other hand, in the frequency domain analysis, particular attention is devoted to the low- and mid-frequency components that encapsulate both global structural patterns and contour characteristics, along with meso-scale features encompassing edge delineation and textural variations. High-frequency components containing fine-grained details and rapidly changing features such as intricate textures and micro-details, introduce noise that can adversely affect depth prediction accuracy. The Deep Frequency Signal Purification (DFSP) module that leverages the Inverse Fourier Transform Theorem \cite{discrete, fourier} to extract global features in the frequency domain, while performing feature purification operations within this spectral space to maximize retention of critical features and eliminate noise interference. Our approach enables effective reconstruction of processed time-domain signals from purified frequency representations while significantly reducing computational complexity compared to the original self-attention mechanism for global information processing, achieving memory complexity reduction from $\mathcal{O}(hN^2+Nd)$ to $\mathcal{O}(NC)$ and time complexity reduction from $\mathcal{O}(N^2d)$ to $\mathcal{O}(N(C+logN))$, where $h$, $C$ denote the attention heads count and feature channels.

\subsection{Interaction and Enhancement of Local Features}

\textbf{Shuffle-Dilation Convolution (SDC).}
The dilated convolution offers the advantage of expanding the receptive field without introducing additional training parameters.  Lite-Mono \cite{litemono} employs stacked dilated convolution layers to capture broader contextual information, but this architectural choice introduces expanded intervals between sampling points, potentially compromising local feature continuity and overlooking fine-grained details. Our proposed Shuffle-Dilation Convolution (SDC) module addresses this limitation by enabling effective utilization of expanded feature channels while reducing the number of dilated convolution layers.  This design preserves local structural continuity while enhancing cross-channel information exchange, as illustrated in Figure {\color{red} 3(a)}.  Formally, given input features $ X \in R^{H \times W \times C}$ and the global channel shuffle operation $Shu_G$, the output $\hat{X}$ generated by our SDC module can be expressed as:
\begin{equation}
    \begin{aligned}
      \hat{X} &= Shu_G[Shu_1(X), X'],
    \end{aligned}
\end{equation}
\begin{equation}
    \begin{aligned}
      F_r(X) &= Linear(LR(Norm(DConv_r(X)))), 
    \end{aligned}
\end{equation}
\begin{equation}
    \begin{aligned}
      X' &=  F_r \circ F_{r-1} \circ \cdots \circ F_1(Shu_2(X))+Shu_2(X) , 
    \end{aligned}
\end{equation}
where $Shu_i$ denotes the group channel shuffle, $[\cdot]$ represents the concatenation operation, $LR$ is the LeakyReLU \cite{leaky} activation function, $Norm$ means LayerNorm and BatchNorm, and $DConv_r$ is the dilated convolution with kerner size 3 and dilation $r$, $\circ$ represents the composition operator.

\textbf{Rotation-Adaptive Kernel Attention (RAKA).}
As an intermediate stage within the encoder architecture, this module serves dual purposes: enhancing post-interaction local features and facilitating global feature extraction. It establishes interdimensional dependencies through a combination of rotation operations and residual connections. This design philosophy effectively mitigates the influence of indirect correspondence between channels and weighting parameters. Specifically, the architecture employs attention gates to dynamically compute dimension-specific attention weights across spatial domains (channel, width, and height) induced by rotational operations. Subsequently, an adaptive convolution kernel scaling mechanism is employed to accommodate multi-scale features. Finally, the linearly combined inverse-rotated tensor is output. The overall process is depicted in Figure {\color{red} 3(b)}. Assuming that the input feature is $X$, we use the following formulas to compute enhanced feature $X^*$: 
\begin{equation}
    \begin{aligned}
      x_i &= Rotation_{i}(X), 
    \end{aligned}
\end{equation}
\begin{equation}
    \begin{aligned}
      x_{pool} = [MaxPool(x_i), AvgPool(x_i)], 
    \end{aligned}
\end{equation}
\begin{equation}
    \begin{aligned}
      X^* &= \sum_{i}  \lambda_i (Rotation^{-1}_{i} (\sigma(Conv_k(x_{pool}))*x_i)), 
    \end{aligned}
\end{equation}
where $x_i$ is the rotation tensor, $\lambda_i$ means the weights assigned by different $x_i$, $\sigma$ denotes the sigmoid activation function, $Conv_k$ is the adaptive kernel convolution, and $Rotation^{-1}_{i}$ is the inverse rotation corresponding to different rotation modes.

\begin{figure*}[h]
\centering
\begin{subfigure}[t]{0.4\textwidth}
  \centering
  \includegraphics[width=\linewidth]{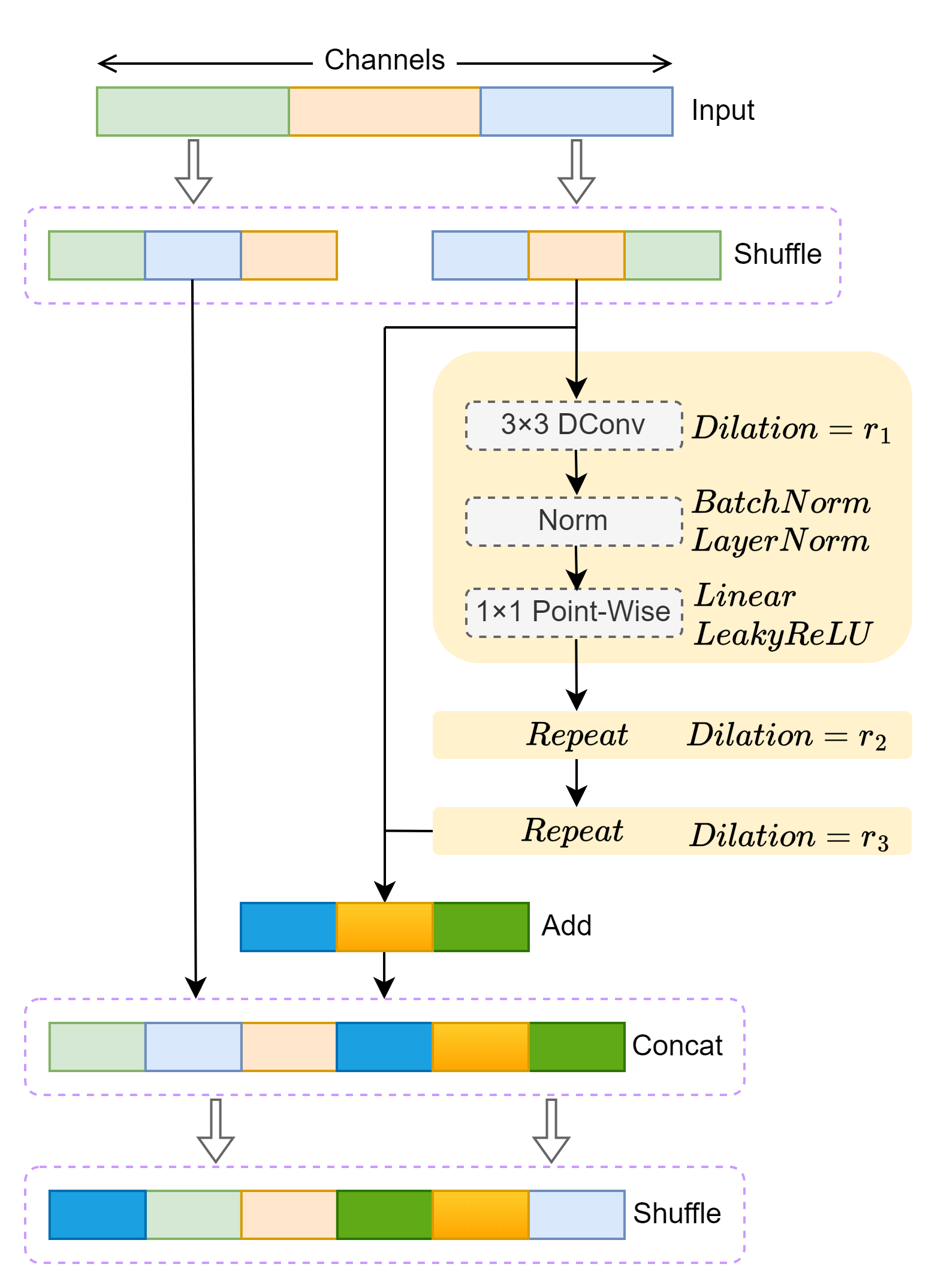} 
  \caption{SDC Module}
  \label{fig:SDC}
\end{subfigure}%
\hfill
\raisebox{0pt}[0pt][0pt]{%
  \begin{subfigure}[t]{0.45\textwidth}
    \centering
    \vspace*{-9.5cm} 
    \includegraphics[width=\linewidth]{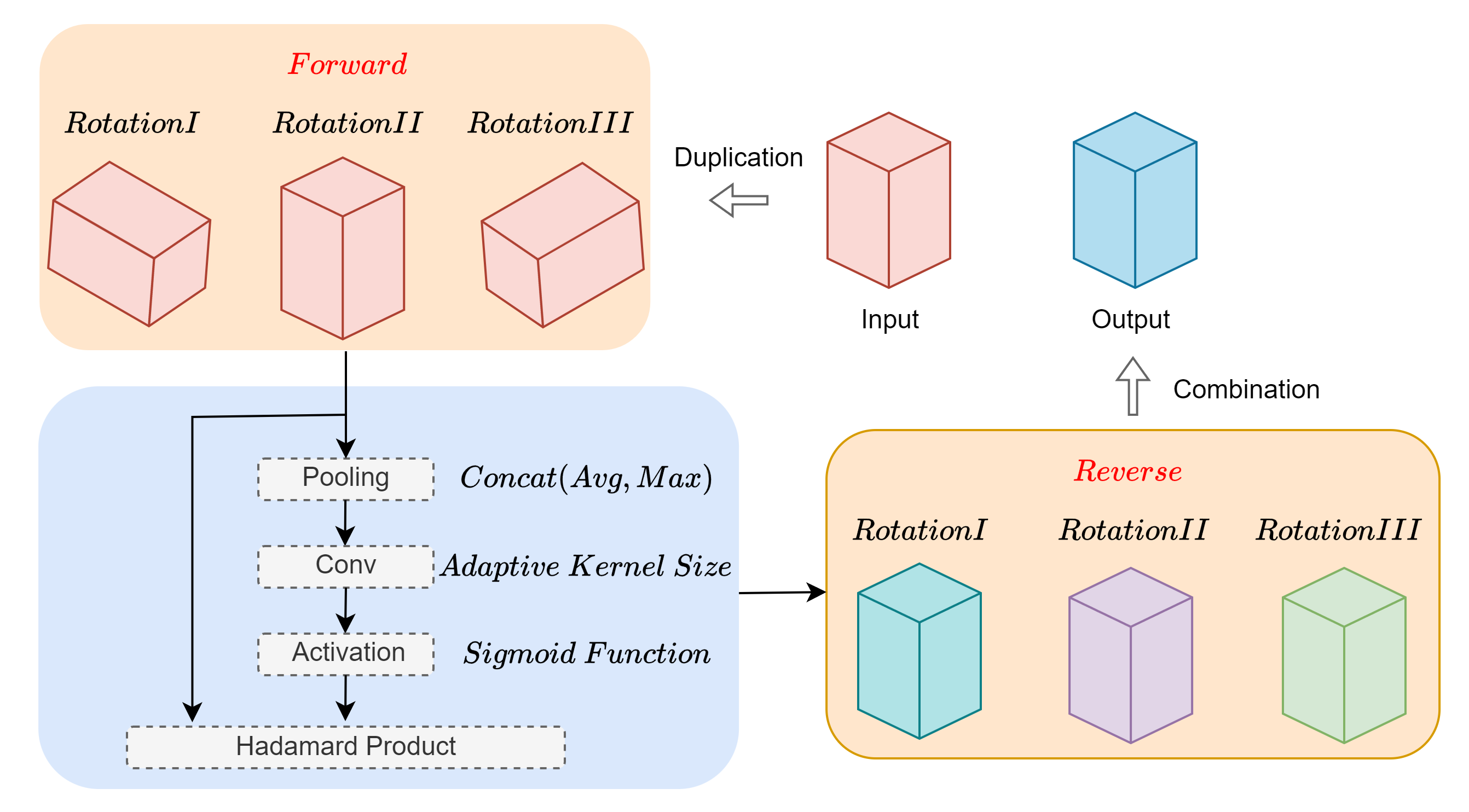} 
    \caption{RAKA Module}
    \label{fig:RAKA}
    \vspace{-0.2cm}
    \includegraphics[width=\linewidth]{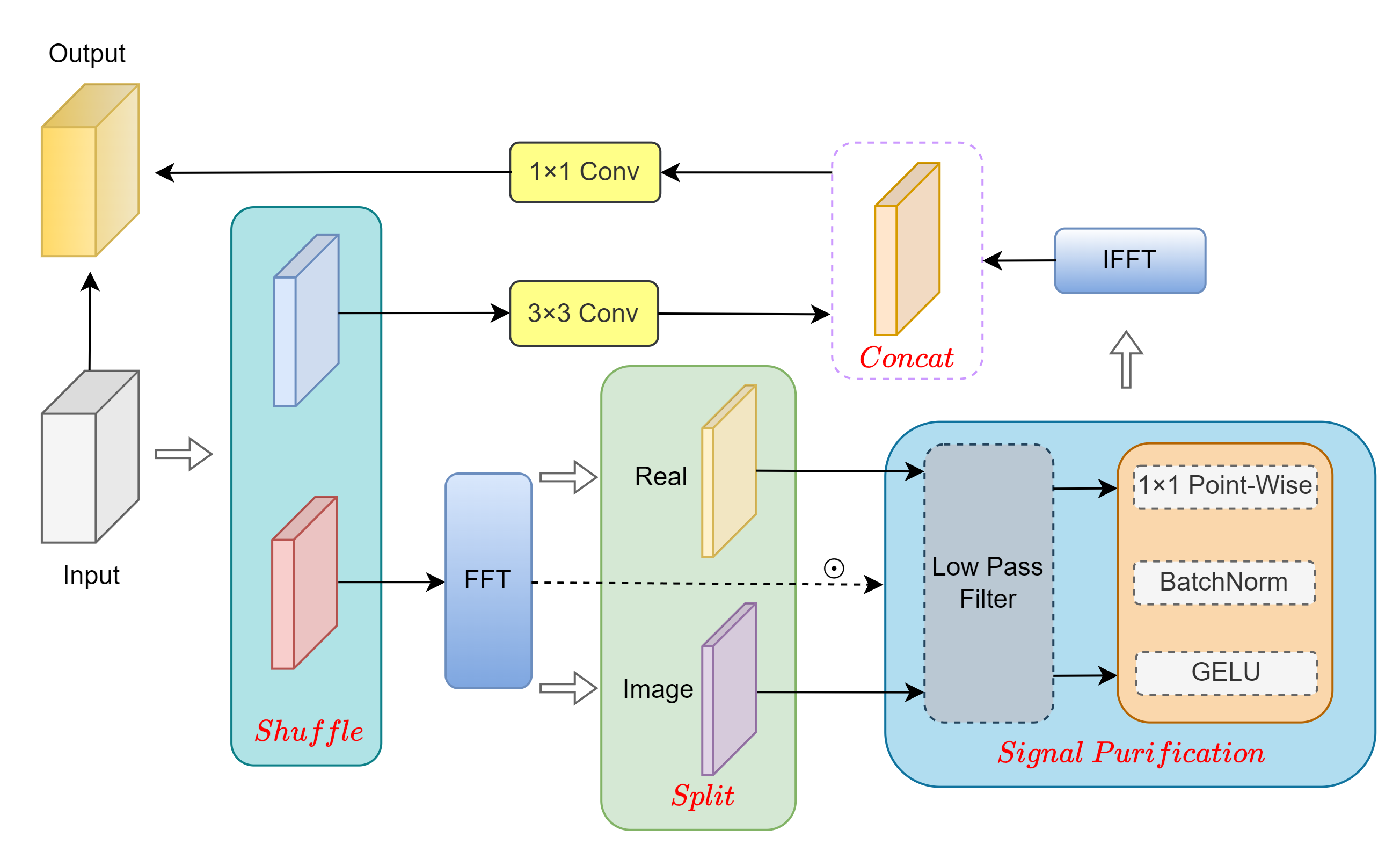} 
    \caption{DFSP Module}
    \label{fig:DFSP}
  \end{subfigure}%
}
\vspace{0.5cm}
\caption{Structures of the proposed SDC, RAKA and DFSP module.}
\label{fig:combined}
\vspace{0.5cm}
\end{figure*}

\subsection{Purified Processing of Global Features}
\textbf{Deep Frequency Signal Purification (DFSP).} The DFSP module effectively purifies unnecessary information while maintaining sufficient global context extraction with minimal parameter overhead.  It leverages the Inverse Fourier Transform Theorem, which states that the Fourier Transform of the convolution operation between two functions equals the Hadamard product in the corresponding Fourier domain.  This mathematical property proves particularly advantageous for resource-constrained applications. Given an enhanced feature $X$, we first apply the Fast Fourier Transform (FFT) to get the frequency representation $X_{fre}$. The operational workflow of DFSP is depicted in Figure {\color{red} 3(c)}, with its output expressed as:
\begin{equation}
    \begin{aligned}
      X_{out} = Conv_p[Conv(Shu_1(X)),\tilde{X}],
    \end{aligned}
\end{equation}
\begin{equation}
    \begin{aligned}
      \tilde{X} = \mathcal{F}^{-1}\ ((\mathcal{P} \circ \mathcal{F}(Shu_2(X))) \odot \mathcal{F}(Shu_2(X))),
    \end{aligned}
\end{equation}
where $Conv_p$ means point-wise convolution, $\mathcal{F}^{-1}$ represents Inverse Fourier Transform (IFT), $\mathcal{P}$ is a purification operator. 

We then introduce the purification operator $\mathcal{P}$. The process begins by decoupling the input $X_{fre}$ into real and imaginary components. Subsequently, a corresponding learnable low-pass filter mask $M$ with the same dimensions as $X_{fre}$ is generated. The frequency radius of the low-pass filter mask $M$ is defined as $\rho(u,v)=\sqrt{u^2+v^2}$, where $u$ and $v$ represent normalized frequency coordinates. The mathematical formulation of the mask $M$ is expressed as:
\begin{equation}
	M(u,v)=
\begin{cases}
	1, $\quad\quad$ \rho(u,v) \leqslant \gamma \cdot max(\rho)\\
	0, $\quad\quad\quad\quad$ otherwise
	\label{dot_hst12}
\end{cases}	
\end{equation}
where $\gamma$ denotes the cutoff ratio, and $max(\rho)$ represents the maximum radius in the frequency domain. These learnable masks adaptively capture multi-frequency information while suppressing undesired high-frequency components, thereby preserving more valuable structural features including shape characteristics and contour details. The frequency-domain filtering operation is then performed by separately applying the mask to the real component $Re(x)$ and imaginary component $Im(x)$ of the input complex tensor $X_{fre}$. Following this, the GELU activation function, convolution and batch normalization operations are applied to the filtered real and imaginary components. Finally, the processed two components are combined to reconstruct the complex representation. The complete process is illustrated in the following equation:
\begin{equation}
    \begin{aligned}
      \mathcal{P}(x)=&Conv_1(BN_{real}(GELU(Re(x)\odot M)))\\
              &+j \cdot Conv_1(BN_{imag}(GELU(Im(x)\odot M))).
    \end{aligned}
\end{equation}
Then an Inverse Fourier Transform is applied to convert the frequency-domain features back to the spatial domain. This frequency-domain operational mechanism effectively achieves effects comparable to spatial-domain convolutions with expanded receptive fields. Equation (8) mathematically establishes the equivalence of employing large dynamic convolution kernels. Based on the Inverse Fourier Transform Theorem stated above, we can formally prove this conclusion:
\begin{equation}
    \begin{aligned}
      \tilde{X} &= \mathcal{F}^{-1}\ ((\mathcal{P} \circ \mathcal{F}(X_s)) \odot \mathcal{F}(X_s)) \\
      &= \mathcal{F}^{-1}\{\mathcal{F} \{ \mathcal{F}^{-1} ((\mathcal{P} \circ \mathcal{F}(X_s)))*X_s\}\} \\
      &= \mathcal{F}^{-1} ((\mathcal{P} \circ \mathcal{F}(X_s)))*X_s,
    \end{aligned}
\end{equation}
where $X_{s}$ is $Shu_2(X)$, $\mathcal{F}^{-1} ((\mathcal{P} \circ \mathcal{F}(X_s)))$ can be viewed spatially as a dynamic convolution kernel of the same size as $X_{s}$. $*$ is spatial-domain convolution.

Note the final equality is approximate, not exact. This approximation stems from the selective filtering and purification of global frequency components during frequency domain processing, introducing inherent cross-domain information loss. We address this issue through a learnable loss weighting mechanism that adaptively adjusts compensation coefficients during optimization.

\subsection{The Decoder of PuriLight}
As evidenced in \cite{litemono}, a simple yet effective decoder architecture demonstrates remarkable suitability for lightweight monocular depth estimation tasks. Following this principle, our architectural decisions consciously prioritize computational efficiency and structural simplicity over sophisticated module integration. We avoid integrating additional components, such as attention mechanisms \cite{bae2022monoformer}. Instead, we adhere to the design philosophy established in \cite{godard2019digging}. Specifically, the up-sampling module employs convolutional blocks to directly concatenate multi-scale features from different hierarchical levels, followed by bilinear upsampling. The final depth maps are generated through a sigmoid activation function, producing outputs at full, 1/2, and 1/4 of the input resolution, respectively.

\subsection{Pose Network and Self-Supervised Training}

\begin{figure}[ht] 
  \centering 
  \includegraphics[width=0.48\textwidth]{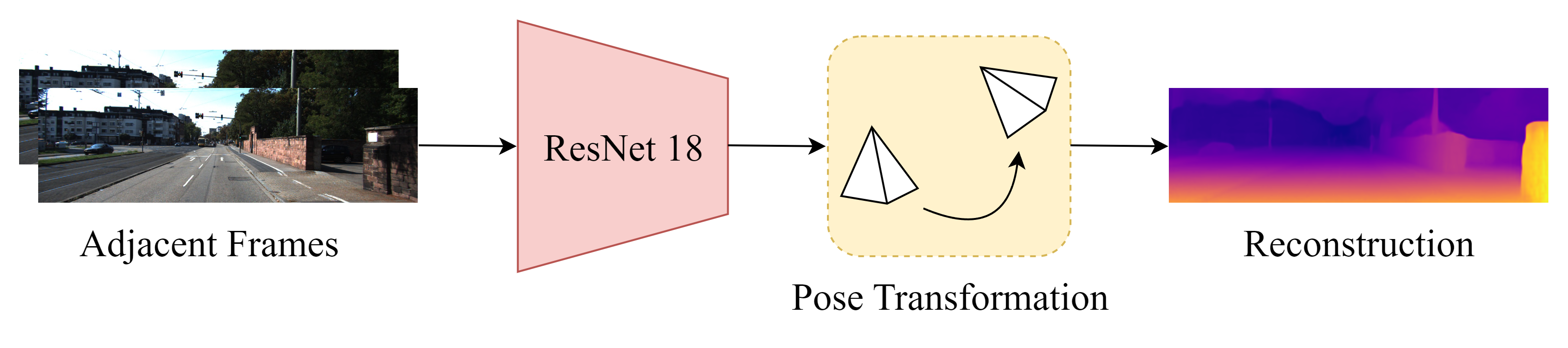} 
  \captionsetup{justification=centering}
  \caption{PoseNet is employed to estimate the poses between adjacent frames. }
  \vspace{0.5cm}
  \label{fig:example pose} 
\end{figure}

As shown in Figure {\color{red} 4}, this work treats monocular depth estimation as an image reconstruction task. Specifically, we approach this problem by formulating it as minimizing the photometric reprojection error during the training process. Following \cite{godard2019digging}, our implementation employs an identical PoseNet architecture for camera pose estimation. The network receives a pair of consecutive RGB frames as input, which are processed through a pose encoder composed of ResNet-18 layers. The architecture subsequently estimates the relative pose $T_{t\to{t'}}$ between the target image $I_{t}$ and source image $I_{t'}$ via a pose decoder consisting of four convolutional layers. Based on this estimated relative pose, we predict a dense depth map $D_{t}$ by optimizing the photometric reprojection loss $L_{p}$, which is defined as:
\begin{equation}
    \begin{aligned}
      L_{p}(I_{t'}, I_{t}) &= \sum_{t'} {pe}(I_{t'\to{t}}, I_{t}) \\ 
      &= \sum_{t'} {pe}(I_{t'} \big \langle  proj(D_{t}, T_{t\to{t'}}, K) \big \rangle, I_{t}) , 
    \end{aligned}
\end{equation}
where $pe$ denotes the photometric reconstruction error, $\langle \cdot  \rangle$ means the sampling operator, ${proj}(\cdot)$ represents the two-dimensional coordinate of $D_{t}$ in $I_{t'}$ and $K$ is the camera’s intrinsics. We use the design in \cite{zhou2017un} where $L_{p}$ consists of $SSIM$ \cite{ssim} and $L_1$ loss:
\begin{equation}
    \begin{aligned}
      pe(I_{a}, I_{b}) = \frac{\alpha}{2} (1-SSIM(I_{a}, I_{b}))+(1-\alpha) \|I_{a}-I_{b}\|,
    \end{aligned}
\end{equation}
where $\alpha$ is empirically set at 0.85 through experiments \cite{godard2019digging} and $\|\cdot\|$ means the $L_1$ norm. To address the challenges posed by out-of-view pixels resulting from self-motion estimation at image boundaries and occluded regions within the visual field, we employ the minimum photometric loss computation method proposed in \cite{godard2019digging}:
\begin{equation}
    \begin{aligned}
      L_{p}(I_{s}, I_{t}) = \mathop{min}_{I_{s}\in[-1, 1]} L_{p}(I_{t'}, I_{t}),
    \end{aligned}
\end{equation}
where $I_{s}\in[-1, 1]$ represents the previous frame and the next frame relative to the target image $I_{t}$. Then, the problem caused by moving pixels is alleviated by a binary mask $\mu(\cdot) \in \{0,1\}$ \cite{godard2019digging}:
\begin{equation}
    \begin{aligned}
      \mu = \big \lbrack \mathop{min}_{I_{s}\in[-1, 1]} L_{p}(I_{s}, I_{t})>\mathop{min}_{I_{s}\in[-1, 1]} L_{p}(I_{t'}, I_{t})  ~ \big \rbrack,
    \end{aligned}
\end{equation}
where $ \lbrack \cdot \rbrack$ is the Iverson bracket. According to \cite{godard2019digging}, an additional edge-aware smoothness loss $L_{s}$ is introduced to estimate depth:
\begin{equation}
    \begin{aligned}
      L_{s} =  \arrowvert\partial_{x} {d_{t}}^*\arrowvert e^{-\arrowvert\partial_{x}I_t\arrowvert} + \arrowvert\partial_{y} {d_{t}}^*\arrowvert e^{-\arrowvert\partial_{y}I_t\arrowvert} ,
    \end{aligned}
\end{equation}
where ${d_t}^* = d_t/\overline {d_t} $ represents the mean-normalized inverse depth to avoid near-zero values and ensure numerical stability during training. Final loss $L_{final}$ combines $L_p$ and $L_s$, as well as the previously described $\mu$ and $\lambda$ of value 1$e^{-3}$:
\begin{equation}
    \begin{aligned}
      L_{final} =  \mu L_{p} + \lambda L_{s},
    \end{aligned}
\end{equation}
and average values are taken on each pixel, scale, and batch.


\section{Experiments}
\label{sec:result}

This section conducts experiments to validate our proposed architecture, showing that PuriLight is an effective lightweight architecture and demonstrates its superiority.

\begin{table*}[h]
\centering
\captionsetup{justification=centering}
\caption{\textbf{Quantitative results.} We compare PuriLight with representative methods on the KITTI Eigen split \cite{eig2015pr}. The best and second-best results are highlighted in \textbf{bold} and \underline{underlined}, respectively. M$^\dagger$: not pre-trained on ImageNet \cite{imagenet}, M: KITTI monocular videos, MS: monocular videos + semantic segmentation, M$^*$: input resolution of 1024 × 320.}
\footnotesize
\setlength{\tabcolsep}{3pt} 
\resizebox{0.9\textwidth}{!}{%
\begin{tabular}{cccccccccc}
\toprule
\multirow{2}{*}{Method} & \multirow{2}{*}{Train} &
\multicolumn{4}{c}{\textbf{Depth Error} $\downarrow$} &
\multicolumn{3}{c}{\textbf{Depth Accuracy} $\uparrow$} &
\multirow{2}{*}{\begin{tabular}[c]{@{}c@{}} \textbf{Model Size} $\downarrow$\\ Params.\end{tabular}} \\
 & & Abs Rel & Sq Rel & RMSE & RMSE log & $\delta < 1.25$ & $\delta < 1.25^2$ & $\delta < 1.25^3$ & \\
\midrule

Monodepth2-Res18 \cite{godard2019digging} & M$^*$ & 0.115   & 0.882  & 4.701 & 0.190     & 0.879 & 0.961 & 0.982 & 14.3M  \\ 
R-MSFM3 \cite{rmsfm}     & M$^*$     & 0.112   & 0.773  & 4.581 & 0.189     & 0.879 & 0.960 & 0.982 & 3.5M   \\  
R-MSFM6 \cite{rmsfm}     & M$^*$     & 0.108   & 0.748  & 4.470 & 0.185     & 0.889 & 0.963 & 0.982 & 3.8M   \\ 
HR-Depth \cite{lyu2021hr}    & M$^*$     & 0.106   & 0.755  & 4.472 & 0.181     & 0.892 & \underline{0.966} & \underline{0.984} & 14.7M  \\ 
Lite-Mono \cite{litemono}   & M$^*$    & \underline{0.102} & \underline{0.746} & \underline{4.444} & \underline{0.179} & \underline{0.896} & 0.965 & 0.983  & \underline{3.1M}  \\ 
PuriLight(Ours)      & M$^*$      & \textbf{0.100}    & \textbf{0.718}   & \textbf{4.403}  & \textbf{0.177}   & \textbf{0.901}   & \textbf{0.967}   & \textbf{0.985} &  \textbf{2.7M}   \\
\midrule

GeoNet \cite{yin2018geonet}  & M & 0.149   & 1.060  & 5.567 & 0.226     & 0.796 & 0.935 & 0.975 & 31.6M  \\ 
DDVO \cite{wang2018} & M & 0.151   & 1.257  & 5.583 & 0.228   & 0.810 &  0.936 & 0.974 & 28.1M  \\ 
Monodepth2-Res18 \cite{godard2019digging} & M & 0.115   & 0.903  & 4.863 & 0.193     & 0.877 & 0.959 & 0.981 & 14.3M  \\ 
Monodepth2-Res50 \cite{godard2019digging} & M & 0.110   & 0.831  & 4.642 & 0.187   & 0.883 &  0.962 & 0.982 & 32.5M  \\  
SGDepth \cite{kli} & MS & 0.113   & 0.835  & 4.693 & 0.191     & 0.879 & 0.961 & 0.981 & 16.3M  \\ 
Johnston et al. \cite{johnston20} & M & 0.111   & 0.941  & 4.817 & 0.189   & 0.885 &  0.961 & 0.981 & 14.3M+  \\  
CADepth-Res18 \cite{yan2021} & M & 0.110   & 0.812  & 4.686 & 0.187     & 0.882 & 0.962 & \underline{0.983} & 18.8M  \\ 
HR-Depth \cite{lyu2021hr} & M & 0.109   & 0.792  & 4.632 & 0.185   & 0.884 &  0.962 & \underline{0.983} & 14.7M  \\ 
Lite-HR-Depth \cite{lyu2021hr} & M & 0.116   & 0.845  & 4.841 & 0.190     & 0.866 & 0.957 & 0.982 & \underline{3.1M}  \\ 
R-MSFM3 \cite{rmsfm} & M & 0.114   & 0.815  & 4.712 & 0.193    & 0.876 &  0.959 & 0.981 & 3.5M   \\ 
R-MSFM6 \cite{rmsfm} & M & 0.112   & 0.806  & 4.704  & 0.191    & 0.878 & 0.960 & 0.981 & 3.8M   \\ 
MonoFormer \cite{bae2022monoformer} & M & 0.108   & 0.806  & 4.594 & 0.184   & 0.884 &  \underline{0.963} & \underline{0.983} & 23.9M+  \\ 
Lite-Mono \cite{litemono}    & M    & \underline{0.107} & \underline{0.765} & \underline{4.561} & \underline{0.183} & \underline{0.886} & \underline{0.963} & \underline{0.983}  & \underline{3.1M}  \\ 
PuriLight(Ours) & M  & \textbf{0.106}   & \textbf{0.747}  & \textbf{4.536} & \textbf{0.182}    & \textbf{0.890} & \textbf{0.964}  & \textbf{0.984} & \textbf{2.7M}   \\

\midrule

Monodepth2-Res18 \cite{godard2019digging} & M$^\dagger$ & 0.132   & 1.044  & 5.142 & 0.210     & 0.845 & 0.948 & 0.977 & 14.3M  \\
Monodepth2-Res50 \cite{godard2019digging} & M$^\dagger$ & 0.131   & 1.023  & 5.046 & 0.206   & 0.849 &  0.951 & \underline{0.979} & 32.5M  \\\
R-MSFM3 \cite{rmsfm}        & M$^\dagger$     & 0.128   & 0.965  & 5.019 & 0.207    & 0.853 &  0.951 & 0.977 & 3.5M   \\
R-MSFM6 \cite{rmsfm}       & M$^\dagger$      & 0.126   & 0.944  & 4.981  & \underline{0.204}    & 0.857 & 0.952 & 0.978 & 3.8M   \\
Lite-Mono \cite{litemono}    & M$^\dagger$   & \textbf{0.121} & \textbf{0.876} & \underline{4.918} & \textbf{0.199} & \underline{0.859} & \underline{0.953} & \textbf{0.98}  & \underline{3.1M}  \\ 
PuriLight(Ours)       & M$^\dagger$       & \textbf{0.121}   & \underline{0.908}  & \textbf{4.908} & \textbf{0.199}    & \textbf{0.862} & \textbf{0.954}  & \textbf{0.98} & \textbf{2.7M}   \\

% ...
\bottomrule
\end{tabular}%
}
\label{tab:depth_estimation}
\end{table*}

\begin{figure*}[h] 
  \centering 
  \includegraphics[width=0.9\textwidth]{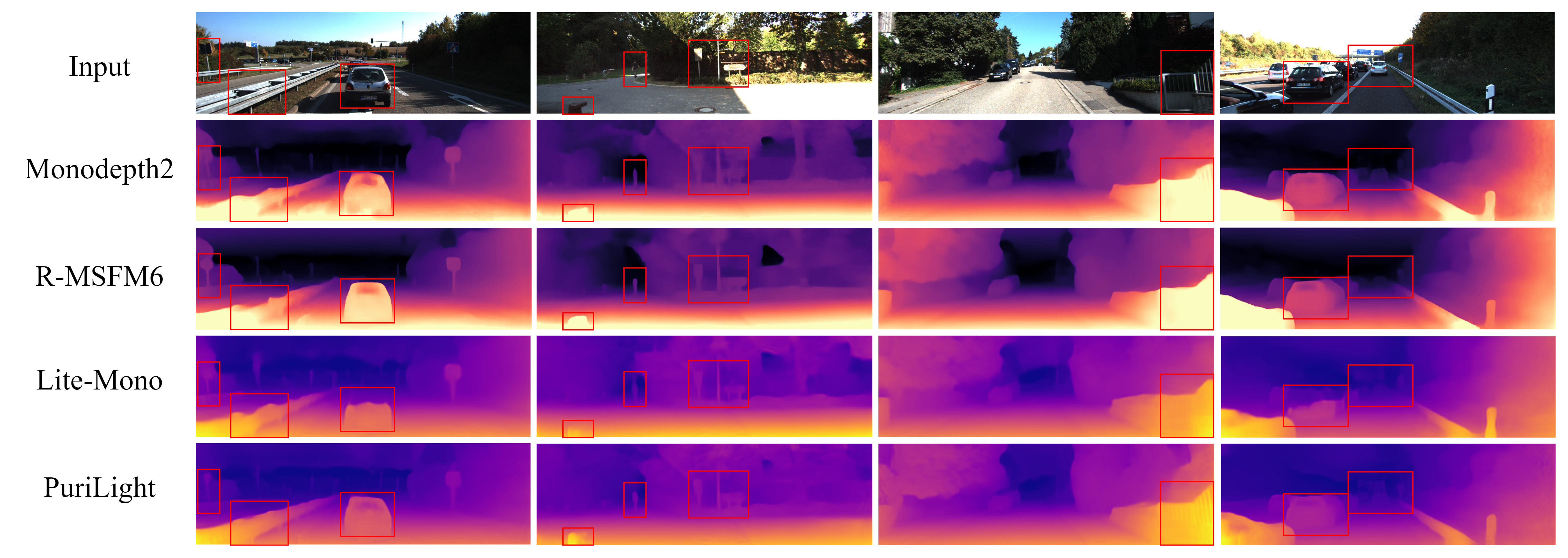} 
  \captionsetup{justification=centering}
  \caption{\textbf{Qualitative results on KITTI}. The proposed PuriLight demonstrates distinct advantages when compared with other representative methods. Monodepth2 \cite{godard2019digging} and R-MSFM \cite{rmsfm} have limited receptive fields, Lite-Mono \cite{litemono} struggles with preserving fine details. In contrast, PuriLight can learn more intricate details and achieve more accurate depth maps.}
  \vspace{0.5cm}
  \label{fig:example3} 
\end{figure*}

\subsection{Specific Details in Training}
The proposed PuriLight framework is implemented using PyTorch and trained on a single NVIDIA GeForce RTX 4090 GPU with a batch size of 16. We adopt AdamW \cite{loshchilov2017decoupled} as the optimizer with weight decay set to 0.01. The training protocol employs a cosine learning rate scheduler \cite{sgdr} initialized at 8$e^{-6}$. Since the pre-training method can greatly increase the convergence speed, the model is trained for 30 epochs when using the pre-trained weights and 60 epochs otherwise. To mitigate overfitting in the proposed method, we implement the following data augmentation strategies with 50\% application probability: horizontal flips, contrast adjustment (±0.2), brightness adjustment (±0.2), random saturation(±0.2), and hue jitter (±0.1). Dropout regularization is additionally employed in SDC and DFSP modules to further address overfitting.

\subsection{Selected Datasets and Experiment Results}

\textbf{KITTI dataset} \cite{kitti} represents a crucial public dataset for autonomous driving research, supporting fundamental computer vision tasks including 3D reconstruction, object detection, tracking, and scene understanding. It provides high-resolution RGB and grayscale images, dense LiDAR point clouds, and accurate pose annotations. We adopt the Eigen split \cite{eig2015pr} for training and evaluation. During the quantitative assessment we restrict the predicted depth values to the 0-80 meter. 

Experiment results on the KITTI dataset are presented in Table {\color{red} 1} and Figure {\color{red}5}, comparing our method with representative approaches.  PuriLight achieves state-of-the-art performance while maintaining the smallest training parameters among baseline models.  Specifically,  compared to the ResNet-50 version of Monodepth2 (32.5M), our method demonstrates superior performance across all evaluation metrics with only 2.7M parameters. This corresponds to a remarkable 91.6\% reduction in parameters. Comparative experiments demonstrate that PuriLight achieves state-of-the-art performance across almost every evaluation metrics compared to other lightweight self-supervised monocular depth estimation approaches, including R-MSFM and Lite-Mono. As shown in Figure {\color{red}5} for qualitative comparisons, PuriLight demonstrates a wider receptive field and more refined structural details. The evaluation metrics adopted are the seven commonly used indicators proposed in \cite{eigen2014depth}, namely Abs Rel, Sq Rel, RMSE, RMSE log, $\delta < 1.25$, $\delta < 1.25^2$, and $\delta < 1.25^3$.

\textbf{Make3D dataset} \cite{make3D} is a public dataset primarily used for 3D scene reconstruction and depth estimation tasks. It contains 134 test images of outdoor scenes, each accompanied by a sparse depth map.  The proposed model is trained on the KITTI dataset and directly evaluated on this test set to validate its generalization capability. As quantitatively demonstrated in Table {\color{red} 2}, PuriLight outperforms four representative baseline methods across all comparative metrics, achieving the most competitive results in cross-dataset evaluation. Qualitative results are presented in Figure {\color{red}6}. It can be observed that in challenging scenarios, our method achieves finer structural details compared to prior lightweight monocular depth estimation models, demonstrating superior generalization performance.
\begin{table}[htbp]
\centering
\captionsetup{justification=centering}
\caption{\textbf{Quantitative results.} Comparison of PuriLight with representative methods on the Make3D \cite{make3D} dataset.}
\resizebox{0.48\textwidth}{!}{%
\begin{tabular}{cccccc}
\toprule
Method     & Abs Rel$\downarrow$ & Sq Rel$\downarrow$ & RMSE$\downarrow$  & RMSE log$\downarrow$ & Params$\downarrow$ \\ \midrule
DDVO \cite{wang2018}      & 0.387   & 4.720  & 8.090 & 0.204    & 28.1M  \\ 
Monodepth2 \cite{godard2019digging} & 0.322   & 3.589  & 7.417 & 0.163    & 14.3M  \\ 
R-MSFM6 \cite{rmsfm}    & 0.334   & 3.285  & 7.212 & 0.169    & 3.8M   \\ 
Lite-Mono \cite{litemono}  & 0.305   & 3.060  & 6.981 & 0.158    & 3.1M   \\ 
PuriLight(Ours)      & \textbf{0.299}   & \textbf{2.922}  & \textbf{6.973} & \textbf{0.157}    & \textbf{2.7M}   \\ 
\bottomrule
\end{tabular}%
}
\label{tab:quantitative}
\end{table}

\begin{table*}[h]
\centering
\captionsetup{justification=centering}
\caption{\textbf{Ablation study of PuriLight architecture.} Models are trained and tested on KITTI.}
\resizebox{0.8\textwidth}{!}{%
\begin{tabular}{cccccccccl}
\toprule
Architecture    & Abs Rel$\downarrow$ & Sq Rel$\downarrow$ & RMSE$\downarrow$  & RMSE log$\downarrow$ & $\delta<1.25$$\uparrow$ & $\delta<1.25^2$$\uparrow$ & $\delta<1.25^3$$\uparrow$ & Params$\downarrow$ & FLOPs$\downarrow$  \\ 
\midrule
adjust SDC blocks  & 0.111        & 0.849       & 4.694      & 0.186         & 0.883      & 0.961      & 0.981      & 2.696M       & 7.105G       \\ 
without RAKA blocks &  0.107       & 0.775       & 4.556      & 0.183         & 0.888      & 0.963      & 0.983      & 2.695M       & 7.095G       \\ 
without DFSP blocks &  0.108       &  0.815      &  4.634     &  0.184        & 0.886      &  0.962     & 0.982      & 2.571M       & 7.044G       \\ 
PuriLight full model      & \textbf{0.106}   & \textbf{0.747}  & \textbf{4.536} & \textbf{0.182}    & \textbf{0.890} & \textbf{0.964} & \textbf{0.984} & 2.696M & 7.105G \\ 
\bottomrule
\vspace{0.5cm}
\end{tabular}%
}
\label{tab:ablation}
\end{table*}

\begin{table}[H]
\centering
\captionsetup{justification=centering}
\caption{\textbf{Model complexity and inference speed.} The input resolution is 640×192, and the batch size is set to 16. Params: training parameters, FLOPs: floating point of operations, Speed: inference speed.}
\resizebox{0.48\textwidth}{!}{%
\begin{tabular}{ccccc}  
\toprule  
\multirow{2}{*}{Method} & \multicolumn{2}{c}{Full Model$\downarrow$}                 & \multicolumn{2}{c}{Speed(ms)$\downarrow$}               \\ 
     & \multicolumn{1}{c}{Params(M)} & FLOPs(G) & \multicolumn{1}{c}{Titan XP} & Jetson Xavier \\ 
\midrule
R-MSFM3 \cite{rmsfm}                 &  \multicolumn{1}{c}{3.5}       & 16.5     & \multicolumn{1}{c}{7.8}      & 22.3          \\ 
R-MSFM6  \cite{rmsfm}                &  \multicolumn{1}{c}{3.8}       & 31.2     & \multicolumn{1}{c}{13.1}     & 41.7          \\ 
MonoViT-tiny  \cite{bae2022monoformer}           &  \multicolumn{1}{c}{10.3}      & 23.7     & \multicolumn{1}{c}{13.5}     & 47.4          \\ 
Lite-Mono-8M \cite{litemono}            &  \multicolumn{1}{c}{8.7}       & 11.2     & \multicolumn{1}{c}{6.5}      & 32.2          \\ 
PuriLight(Ours)                    &  \multicolumn{1}{c}{\textbf{2.7}}       & \textbf{7.1}      & \multicolumn{1}{c}{\textbf{4.5}}      & \textbf{20.2}          \\ 
\bottomrule
\end{tabular}%
}
\label{tab:complexity}
\end{table}

\subsection{Ablation Experiment of Model Structure}
This section shows the importance of conducting ablation experiments on the proposed framework to verify different modules and further prove the effectiveness of the proposed model. We delete or adjust the modules of the experiment, and conduct training on the KITTI dataset in the same way as the original experiment. The quantitative outcomes are presented in Table {\color{red} 3}. It can be observed that adjusting or removing the proposed modules results in varying degrees of performance degradation compared to the full model, which validates the utility of our design. We further demonstrate adjustment strategies for distinct modules and their advantages.

(i) \textbf{SDC modules.} We remove the shuffle operations introduced at all stages of the model and set the dilation rates of the dilated convolutions to 1 uniformly. Experimental results demonstrate that while the model size remains unchanged, a significant performance degradation decrease is observed. This observation highlights that the benefits of employing SDC modules lie in their ability to progressively extract multi-scale features for enhanced local feature representation and facilitate inter-channel information exchange, both of which prove crucial for maintaining model efficacy.

(ii) \textbf{RAKA modules.} Following the removal of RAKA modules across all network stages, we observe a few reductions in both parameters and key performance metrics. This observation indicates that the hierarchical architecture of RAKA modules enhances the expressive capacity of multi-scale feature representations and facilitates subsequent processing by downstream modules through optimized feature alignment.

(iii) \textbf{DFSP modules.} When the DFSP modules are removed from the architecture, experimental results reveal that although the model parameters experience a slight reduction, there is a moderate degradation in performance metrics. This observation suggests that the DFSP modules not only effectively capture global information patterns but also perform critical noise suppression by filtering out non-essential spectral features and environmental artifacts.

\subsection{Model Complexity and Inference Speed}

We conduct comprehensive evaluations of the proposed model's parameters, FLOPs (floating-point operations), and inference time on both NVIDIA TITAN Xp and Jetson Xavier platforms, with comparative analysis against several state-of-the-art approaches. As summarized in Table {\color{red} 4}, our design achieves a favorable balance between model complexity and inference speed. Furthermore, the proposed architecture maintains real-time inference capability on Jetson Xavier, indicating its practical applicability for edge computing deployments.


\section{Conclusion}
\label{sec:conclusion}
We propose PuriLight, a novel lightweight self-supervised monocular depth estimation framework. It systematically processes input data through three dedicated modules: a local feature extractor for detail preservation, a feature enhancement unit for contextual reinforcement, and a global feature purification mechanism. Comprehensive evaluations on the KITTI dataset demonstrate that our model achieves state-of-the-art performance with the fewest parameters. Notably, cross-dataset validation on Make3D reveals superior generalization capability compared to existing approaches. Ablation experiments further verify the architecture's advantages in both computational efficiency and model complexity.

\begin{figure}[htbp] 
  \centering 
  \includegraphics[width=0.4\textwidth]{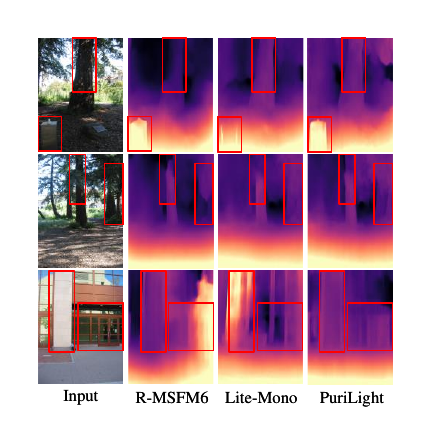} 
  \captionsetup{justification=centering}
  \caption{\textbf{Qualitative results on Make3D}. The performance of the model is validated on challenging scenarios from the Make3D dataset. It can be observed that PuriLight achieves a wider receptive field and finer details compared to prior lightweight models, while demonstrating superior generalization capability.}
  \vspace{0.5cm}
  \label{fig:example4} 
\end{figure}

\section{Limitations}

During the visualization process, we encounter instances where certain images lose partial details, as demonstrated in the upper-right corner of the last image in Figure {\color{red} 6}. This indicates that our model may occasionally misclassify detailed textures and contour features as noise for removal. The phenomenon becomes more pronounced when dynamic objects are present. Future work will focus on developing a lightweight dynamic object removal model to address this limitation.

\clearpage
\titlespacing*{\section}{0pt}{0pt}{10pt}
\bibliography{main}

@inproceedings{eigen2014depth,
 author = {Eigen, David and Puhrsch, Christian and Fergus, Rob},
 booktitle = {Advances in Neural Information Processing Systems},
 pages={2366-2374},
 title = {Depth Map Prediction from a Single Image using a Multi-Scale Deep Network},
 volume={27},
 year = {2014}
}

@INPROCEEDINGS{laina2016deeper,
  author={Laina, Iro and Rupprecht, Christian and Belagiannis, Vasileios and Tombari, Federico and Navab, Nassir},
  booktitle={2016 Fourth International Conference on 3D Vision (3DV)}, 
  title={Deeper Depth Prediction with Fully Convolutional Residual Networks}, 
  year={2016},
  volume={},
  number={},
  pages={239-248},
 }

@INPROCEEDINGS{fu2018deep,
  author={Fu, Huan and Gong, Mingming and Wang, Chaohui and Batmanghelich, Kayhan and Tao, Dacheng},
  booktitle={2018 IEEE/CVF Conference on Computer Vision and Pattern Recognition}, 
  title={Deep Ordinal Regression Network for Monocular Depth Estimation}, 
  year={2018},
  volume={},
  number={},
  pages={2002-2011},
}

@INPROCEEDINGS{yin2018geonet,
  author={Yin, Zhichao and Shi, Jianping},
  booktitle={2018 IEEE/CVF Conference on Computer Vision and Pattern Recognition}, 
  title={GeoNet: Unsupervised Learning of Dense Depth, Optical Flow and Camera Pose}, 
  year={2018},
  volume={},
  number={},
  pages={1983-1992},
}

@INPROCEEDINGS{he2016deep,
  author={He, Kaiming and Zhang, Xiangyu and Ren, Shaoqing and Sun, Jian},
  booktitle={2016 IEEE Conference on Computer Vision and Pattern Recognition (CVPR)}, 
  title={Deep Residual Learning for Image Recognition}, 
  year={2016},
  volume={},
  number={},
  pages={770-778},
}

@INPROCEEDINGS{godard2019digging,
  author={Godard, Clement and Aodha, Oisin Mac and Firman, Michael and Brostow, Gabriel},
  booktitle={2019 IEEE/CVF International Conference on Computer Vision (ICCV)}, 
  title={Digging Into Self-Supervised Monocular Depth Estimation}, 
  year={2019},
  volume={},
  number={},
  pages={3827-3837},
}

@inproceedings{lyu2021hr,
  title={Hr-depth: High resolution self-supervised monocular depth estimation},
  author={Lyu, Xiaoyang and Liu, Liang and Wang, Mengmeng and Kong, Xin and Liu, Lina and Liu, Yong and Chen, Xinxin and Yuan, Yi},
  booktitle={Proceedings of the AAAI conference on artificial intelligence},
  volume={35},
  pages={2294--2301},
  year={2021}
}

@article{zhou2021self,
  title={Self-supervised monocular depth estimation with internal feature fusion},
  author={Zhou, Hang and Greenwood, David and Taylor, Sarah},
  journal={arXiv preprint arXiv:2110.09482},
  year={2021}
}

@INPROCEEDINGS{yang2021trans,
  author={Yang, Guanglei and Tang, Hao and Ding, Mingli and Sebe, Nicu and Ricci, Elisa},
  booktitle={2021 IEEE/CVF International Conference on Computer Vision (ICCV)}, 
  title={Transformer-Based Attention Networks for Continuous Pixel-Wise Prediction}, 
  year={2021},
  volume={},
  number={},
  pages={16249-16259},
}

@INPROCEEDINGS{ran221vis,
  author={Ranftl, René and Bochkovskiy, Alexey and Koltun, Vladlen},
  booktitle={2021 IEEE/CVF International Conference on Computer Vision (ICCV)}, 
  title={Vision Transformers for Dense Prediction}, 
  year={2021},
  volume={},
  number={},
  pages={12159-12168},
}

@INPROCEEDINGS{ke2024rep,
  author={Ke, Bingxin and Obukhov, Anton and Huang, Shengyu and Metzger, Nando and Daudt, Rodrigo Caye and Schindler, Konrad},
  booktitle={2024 IEEE/CVF Conference on Computer Vision and Pattern Recognition (CVPR)}, 
  title={Repurposing Diffusion-Based Image Generators for Monocular Depth Estimation}, 
  year={2024},
  volume={},
  number={},
  pages={9492-9502},
}

@INPROCEEDINGS{chang2018,
  author={Chang, Jia-Ren and Chen, Yong-Sheng},
  booktitle={2018 IEEE/CVF Conference on Computer Vision and Pattern Recognition}, 
  title={Pyramid Stereo Matching Network}, 
  year={2018},
  volume={},
  number={},
  pages={5410-5418},
}

@InProceedings{garg2016,
author="Garg, Ravi
and B.G., Vijay Kumar
and Carneiro, Gustavo
and Reid, Ian",
title="Unsupervised CNN for Single View Depth Estimation: Geometry to the Rescue",
booktitle="Computer Vision -- ECCV 2016",
year="2016",
publisher="Springer International Publishing",
pages="740--756",
}

@INPROCEEDINGS{godard2017,
  author={Godard, Clément and Aodha, Oisin Mac and Brostow, Gabriel J.},
  booktitle={2017 IEEE Conference on Computer Vision and Pattern Recognition (CVPR)}, 
  title={Unsupervised Monocular Depth Estimation with Left-Right Consistency}, 
  year={2017},
  volume={},
  number={},
  pages={6602-6611},
}

@INPROCEEDINGS{zhou2017un,
  author={Zhou, Tinghui and Brown, Matthew and Snavely, Noah and Lowe, David G.},
  booktitle={2017 IEEE Conference on Computer Vision and Pattern Recognition (CVPR)}, 
  title={Unsupervised Learning of Depth and Ego-Motion from Video}, 
  year={2017},
  volume={},
  number={},
  pages={6612-6619},
}

@article{kitti,
author = {A Geiger and P Lenz and C Stiller and R Urtasun},
title ={Vision meets robotics: The KITTI dataset},
journal = {The International Journal of Robotics Research},
volume = {32},
number = {11},
pages = {1231-1237},
year = {2013},
}

@ARTICLE{make3D,
  author={Saxena, Ashutosh and Sun, Min and Ng, Andrew Y.},
  journal={IEEE Transactions on Pattern Analysis and Machine Intelligence}, 
  title={Make3D: Learning 3D Scene Structure from a Single Still Image}, 
  year={2009},
  volume={31},
  number={5},
  pages={824-840},
}

@INPROCEEDINGS{rmsfm,
  author={Zhou, Zhongkai and Fan, Xinnan and Shi, Pengfei and Xin, Yuanxue},
  booktitle={2021 IEEE/CVF International Conference on Computer Vision (ICCV)}, 
  title={R-MSFM: Recurrent Multi-Scale Feature Modulation for Monocular Depth Estimating}, 
  year={2021},
  volume={},
  number={},
  pages={12757-12766},
}

@INPROCEEDINGS{litemono,
  author={Zhang, Ning and Nex, Francesco and Vosselman, George and Kerle, Norman},
  booktitle={2023 IEEE/CVF Conference on Computer Vision and Pattern Recognition (CVPR)}, 
  title={Lite-Mono: A Lightweight CNN and Transformer Architecture for Self-Supervised Monocular Depth Estimation}, 
  year={2023},
  volume={},
  number={},
  pages={18537-18546},
}

@INPROCEEDINGS{patni2024,
  author={Patni, Suraj and Agarwal, Aradhye and Arora, Chetan},
  booktitle={2024 IEEE/CVF Conference on Computer Vision and Pattern Recognition (CVPR)}, 
  title={ECoDepth: Effective Conditioning of Diffusion Models for Monocular Depth Estimation}, 
  year={2024},
  volume={},
  number={},
  pages={28285-28295},
}

@InProceedings{silb2012,
author="Silberman, Nathan
and Hoiem, Derek
and Kohli, Pushmeet
and Fergus, Rob",
title="Indoor Segmentation and Support Inference from RGBD Images",
booktitle="Computer Vision -- ECCV 2012",
year="2012",
publisher="Springer Berlin Heidelberg",
pages="746--760",

}

@INPROCEEDINGS{laina16deep,
  author={Laina, Iro and Rupprecht, Christian and Belagiannis, Vasileios and Tombari, Federico and Navab, Nassir},
  booktitle={2016 Fourth International Conference on 3D Vision (3DV)}, 
  title={Deeper Depth Prediction with Fully Convolutional Residual Networks}, 
  year={2016},
  volume={},
  number={},
  pages={239-248},
}

@INPROCEEDINGS{zhou2017uns,
  author={Zhou, Chao and Zhang, Hong and Shen, Xiaoyong and Jia, Jiaya},
  booktitle={2017 IEEE International Conference on Computer Vision (ICCV)}, 
  title={Unsupervised Learning of Stereo Matching}, 
  year={2017},
  volume={},
  number={},
  pages={1576-1584},
}

@article{yu2016multi,
      title={Multi-Scale Context Aggregation by Dilated Convolutions}, 
      author={Fisher Yu and Vladlen Koltun},
      year={2015},
      journal={arXiv preprint arXiv:1511.07122},
      archivePrefix={arXiv}, 
}

@INPROCEEDINGS{zhang2018shu,
  author={Zhang, Xiangyu and Zhou, Xinyu and Lin, Mengxiao and Sun, Jian},
  booktitle={2018 IEEE/CVF Conference on Computer Vision and Pattern Recognition}, 
  title={ShuffleNet: An Extremely Efficient Convolutional Neural Network for Mobile Devices}, 
  year={2018},
  volume={},
  number={},
  pages={6848-6856},
}

@INPROCEEDINGS{triplet,
  author={Misra, Diganta and Nalamada, Trikay and Arasanipalai, Ajay Uppili and Hou, Qibin},
  booktitle={2021 IEEE Winter Conference on Applications of Computer Vision (WACV)}, 
  title={Rotate to Attend: Convolutional Triplet Attention Module}, 
  year={2021},
  volume={},
  number={},
  pages={3138-3147},
}

@inproceedings{transformer,
 author = {Vaswani, Ashish and Shazeer, Noam and Parmar, Niki and Uszkoreit, Jakob and Jones, Llion and Gomez, Aidan N and Kaiser, \L ukasz and Polosukhin, Illia},
 booktitle = {Advances in Neural Information Processing Systems},
 pages = {},
 title = {Attention is All you Need},
 volume = {30},
 year = {2017}
}

@book{discrete,
author="McGillem, Clare D. and Cooper, George R.",
title="Continuous and discrete signal and system analysis",
publisher="Saunders College Pub.",
year="1991",
}

@book{fourier,
author = {Oppenheim, Alan V. and Schafer, Ronald W.},
title = {Discrete-Time Signal Processing},
year = {2009},
publisher = {Prentice Hall Press},
}

@inproceedings{leaky,
  title={Rectifier nonlinearities improve neural network acoustic models},
  author={Maas, Andrew L and Hannun, Awni Y and Ng, Andrew Y and others},
  booktitle={Proc. icml},
  volume={30},
  pages={3},
  year={2013},
}

@article{bae2022monoformer,
  title={Monoformer: Towards generalization of self-supervised monocular depth estimation with transformers},
  author={Bae, Jinwoo and Moon, Sungho and Im, Sunghoon},
  journal={arXiv preprint arXiv:2205.11083},
  volume={1},
  number={2},
  pages={4},
  year={2022}
}

@article{loshchilov2017decoupled,
  title={Decoupled weight decay regularization},
  author={Loshchilov, Ilya and Hutter, Frank},
  journal={arXiv preprint arXiv:1711.05101},
  year={2017}
}

@inproceedings{sgdr,
  author={Ilya Loshchilov and Frank Hutter},
  title={SGDR: Stochastic Gradient Descent with Warm Restarts},
  year={2017},
  booktitle={ICLR (Poster)},
}

@INPROCEEDINGS{imagenet,
  author={Deng, Jia and Dong, Wei and Socher, Richard and Li, Li-Jia and Kai Li and Li Fei-Fei},
  booktitle={2009 IEEE Conference on Computer Vision and Pattern Recognition}, 
  title={ImageNet: A large-scale hierarchical image database}, 
  year={2009},
  volume={},
  number={},
  pages={248-255},
}

@INPROCEEDINGS{eig2015pr,
  author={Eigen, David and Fergus, Rob},
  booktitle={2015 IEEE International Conference on Computer Vision (ICCV)}, 
  title={Predicting Depth, Surface Normals and Semantic Labels with a Common Multi-scale Convolutional Architecture}, 
  year={2015},
  volume={},
  number={},
  pages={2650-2658},
}

@INPROCEEDINGS{wang2018,
  author={Wang, Chaoyang and Buenaposada, José Miguel and Zhu, Rui and Lucey, Simon},
  booktitle={2018 IEEE/CVF Conference on Computer Vision and Pattern Recognition}, 
  title={Learning Depth from Monocular Videos Using Direct Methods}, 
  year={2018},
  volume={},
  number={},
  pages={2022-2030},
}

@InProceedings{kli,
author="Klingner, Marvin
and Term{\"o}hlen, Jan-Aike
and Mikolajczyk, Jonas
and Fingscheidt, Tim",
title="Self-supervised Monocular Depth Estimation: Solving the Dynamic Object Problem by Semantic Guidance",
booktitle="Computer Vision -- ECCV 2020",
year="2020",
publisher="Springer International Publishing",
pages="582--600",
}

@INPROCEEDINGS{johnston20,
  author={Johnston, Adrian and Carneiro, Gustavo},
  booktitle={2020 IEEE/CVF Conference on Computer Vision and Pattern Recognition (CVPR)}, 
  title={Self-Supervised Monocular Trained Depth Estimation Using Self-Attention and Discrete Disparity Volume}, 
  year={2020},
  volume={},
  number={},
  pages={4755-4764},
}

@INPROCEEDINGS{yan2021,
  author={Yan, Jiaxing and Zhao, Hong and Bu, Penghui and Jin, YuSheng},
  booktitle={2021 International Conference on 3D Vision (3DV)}, 
  title={Channel-Wise Attention-Based Network for Self-Supervised Monocular Depth Estimation}, 
  year={2021},
  volume={},
  number={},
  pages={464-473},
}

@inproceedings{wei2023surrounddepth,
  title={Surrounddepth: Entangling surrounding views for self-supervised multi-camera depth estimation},
  author={Wei, Yi and Zhao, Linqing and Zheng, Wenzhao and Zhu, Zheng and Rao, Yongming and Huang, Guan and Lu, Jiwen and Zhou, Jie},
  booktitle={Conference on Robot Learning},
  pages={539--549},
  year={2023},
}

@ARTICLE{ssim,
  author={Zhou Wang and Bovik, A.C. and Sheikh, H.R. and Simoncelli, E.P.},
  journal={IEEE Transactions on Image Processing}, 
  title={Image quality assessment: from error visibility to structural similarity}, 
  year={2004},
  volume={13},
  pages={600-612}
}

@incollection{lee2024robust,
  title={Robust Monocular Depth Estimation in Adverse Weather Conditions by Unsupervised Domain Adaptation},
  author={Lee, Jihui and Lai-Dang, Quoc-Vinh and Sengar, Neha and Har, Dongsoo},
  booktitle={ECAI 2024},
  pages={601--608},
  year={2024},
  publisher={IOS Press}
}

@inproceedings{liu2024mono,
  title={Mono-ViFI: A unified learning framework for self-supervised single and multi-frame monocular depth estimation},
  author={Liu, Jinfeng and Kong, Lingtong and Li, Bo and Wang, Zerong and Gu, Hong and Chen, Jinwei},
  booktitle={European Conference on Computer Vision},
  pages={90--107},
  year={2024},
  organization={Springer}
}
\end{document}